\documentclass[twoside]{article}

\usepackage[utf8]{inputenc} % allow utf-8 input
\usepackage[T1]{fontenc}    % use 8-bit T1 fonts
\usepackage{url}            % simple URL typesetting
\usepackage{booktabs}       % professional-quality tables
\usepackage{amsfonts}       % blackboard math symbols
\usepackage{nicefrac}       % compact symbols for 1/2, etc.
\usepackage{microtype}      % microtypography
\usepackage{bm}
\usepackage{enumitem}
\usepackage{graphicx}
\usepackage{amsmath}
\usepackage{subfigure}
\usepackage{caption}
\usepackage{algorithm}
\usepackage{algorithmic}
\usepackage[dvipsnames]{xcolor}
\usepackage[colorlinks=true,bookmarksopen,pdfsubject={algorithms},linkcolor={MidnightBlue}, anchorcolor={black}, citecolor={MidnightBlue}, filecolor={magenta}, menucolor={black}, pagecolor={red},backref=none,urlcolor={MidnightBlue}]{hyperref}

\usepackage[accepted]{aistats2024}
\usepackage[round]{natbib}

\begin{document}



\twocolumn[

\aistatstitle{Discriminant Distance-Aware Representation on Deterministic Uncertainty Quantification Methods}

\aistatsauthor{ Jiaxin Zhang \\ 
\And Kamalika Das \\ \And  Sricharan Kumar \\ }

\aistatsaddress{ Intuit AI Research  \\ \texttt{jiaxin\_zhang@intuit.com} 
\And  Intuit AI Research \\ \texttt{kamalika\_das@intuit.com}
\And Intuit AI Research \\ \texttt{sricharan\_kumar@intuit.com}}
]

\begin{abstract}
Uncertainty estimation is a crucial aspect of deploying dependable deep learning models in safety-critical systems. In this study, we introduce a novel and efficient method for deterministic uncertainty estimation called Discriminant Distance-Awareness Representation (DDAR). Our approach involves constructing a DNN model that incorporates a set of prototypes in its latent representations, enabling us to analyze valuable feature information from the input data. By leveraging a distinction maximization layer over optimal trainable prototypes, DDAR can learn a discriminant distance-awareness representation. We demonstrate that DDAR overcomes feature collapse by relaxing the Lipschitz constraint that hinders the practicality of deterministic uncertainty methods (DUMs) architectures. Our experiments show that DDAR is a flexible and architecture-agnostic method that can be easily integrated as a pluggable layer with distance-sensitive metrics, outperforming state-of-the-art uncertainty estimation methods on multiple benchmark problems.
\end{abstract}

\section{Introduction}
\label{sec:intro}
Deep neural network (DNN) models play an important role in many safety-critical tasks, e.g., autonomous driving, or medical diagnosis. A key characteristic shared by these tasks is their risk sensitivity so that a confidently wrong prediction can lead to fatal accidents and misleading decisions. Therefore, it is of utmost importance to develop reliable and efficient uncertainty estimation methods that allow for the safe deployment in large-scale, real-world applications across computer vision and natural language processing \citep{zhang2021modern,tran2022plex,zhang2024quantification}. 

However, naive DNN models do not deliver certainty estimates or suffer from over or under-confidence, i.e. are badly calibrated or assigned with high confidence to out-of-domain (OOD) inputs. This has led to the development of probabilistic approaches for uncertainty estimation in DNN models. Bayesian Neural Networks (BNNs) \citep{osawa2019practical, wenzel2020good} represent the dominant solution for quantifying uncertainty but exactly modeling the full posterior is often computationally intractable, and not scale well to complex tasks \citep{mukhoti2021deep}. Monte Carlo (MC) Dropout \citep{gal2016dropout}, is simple to implement but its uncertainty is not always reliable while requiring multiple forward passes. Deep Ensembles \citep{lakshminarayanan2017simple} involves training multiple deep models from different initializations and a different data set ordering, which outperforms BNN but comes at the high expense of computational cost \citep{ovadia2019can}. A shared characteristic of these approaches is their high computational cost and large memory requirement. Thus, efficient and scalable methods for uncertainty estimation largely remain an open problem \citep{gawlikowski2023survey}. 
\begin{figure*}[h!]
    \centering
    \includegraphics[width=0.9\textwidth]{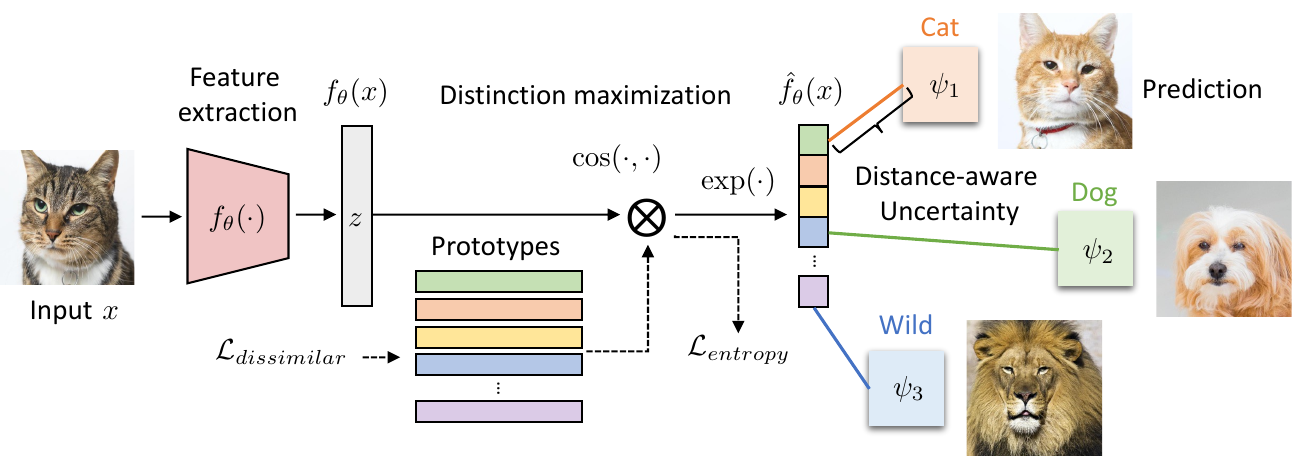}
    \caption{DDAR overview: an efficient, distance-aware and architecture-agnostic method for deterministic uncertainty estimation. DDAR learns a discriminative distance-aware latent representation by leveraging the learnable prototypes and distinction maximization layer. Combined with the RBF kernel, our method performs accurate uncertainty estimation and competitive OOD detection capabilities. }
    \label{fig:overview}
\end{figure*}

% \vspace{-10pt}
Recently, a set of promising works, named Deterministic Uncertainty Methods (DUMs) \citep{mukhoti2021deep} emerged for estimating uncertainty with a single forward pass while treating its weights deterministically \citep{postels2021practicality}. These methods are prone to be efficient and scalable solutions to uncertainty estimation and out-of-distribution (OOD) detection problems. DUMs aim at learning informative latent representation of a model given that the distribution of latent representation should be representative of the input distribution. Then DUMs estimate uncertainty by replacing the final softmax layer with a distance-sensitive function. Specifically, DUQ \citep{van2020uncertainty} defines the uncertainty as the distance between the model output and the closest centroid and proposes a novel centroid updating step based on Radial Basis Function (RBF) networks. DUE \citep{van2021feature} built upon DUQ introduces deep kernel learning by using the inducing point approximation by incorporating a large number of inducing points without overfitting. SNGP \citep{liu2020simple} replaces the softmax layer with Gaussian processes (GP) with RBF kernel to extend distance awareness to the output layer. However, naive latent representations typically suffer from the \textit{feature collapse} \citep{van2020uncertainty} issue when OOD data are mapped to similar feature representations as in-distribution data, which makes OOD detection based on high-level representations impossible. To address the feature collapse issue, DUMs strongly rely on the regularization of latent representation with the ability to differentiate between in-distribution and out-of-distribution data. Otherwise, these methods have several essential challenges and weaknesses \citep{postels2021practicality}. 

Specifically, DUMs mitigate the feature collapse issue through regularization techniques to mimic distances between latent representations to distances in the original input space. This is often achieved by adding constraints over the bi-Lipschitz constant, which enforces a lower and upper bound to expansion and contraction performed by a DNN model \citep{postels2021practicality}. The upper bound enforces \textit{smoothness}, i.e., small changes in the input do not lead to large changes in the latent space and the lower bound enforces \textit{sensitiveness}, i.e., different inputs are mapped to distinct latent spaces. Primarily, there are two methods to impose the bi-Lipschitz constraint: (1) \textit{Gradient Penalty} \citep{van2020uncertainty} directly constrains the gradient of input but leads to large computational cost due to backpropagation through the input's gradients; (2) \textit{Spectral Normalization} \citep{miyato2018spectral} normalizes the weights of each residual layers using their spectral norm, which is computationally more efficient compared with gradient penalty but requires the use of residual layers so it is not architecture-agnostic \citep{van2021feature,liu2020simple}. Moreover, both regularization in distance-awareness representations have limitations in explicitly preserving sample-specific information. In other words, they may discard useful information in the latent representation depending on the underlying distance metric \citep{postels2020hidden, wu2020simple, franchi2022latent}. Although the distance-awareness representation with the above regularization shows promising results, it does not explicitly preserve sample-specific information. In other words, it may discard useful information in its latent representation depending on the underlying distance metric. To fill the gap, this work aims to answer the following questions: 

\begin{itemize}[leftmargin=10pt]
% \vspace{-6pt}
    \item {\em Can we build a simple and efficient uncertainty estimation method without feature collapse issue? }
% \vspace{-3pt}
    \item {\em Is that possible to learn a distance-aware representation while preserving sample-specific information? } 
% \vspace{-3pt}
    \item {\em Is the feature extractor architecture-agnostic, with higher flexibility, not limited by residual layers? }
\end{itemize}
   
To answer these core questions, we develop {DDAR} (Discriminant Distance-Awareness Representation) - a novel method for deterministic uncertainty estimation, which is {\em efficient,  distance-aware, and architecture-agnostic}. As shown in Fig.~\ref{fig:overview}, we first build a DNN model imbued with a set of prototypes over its latent presentations. These prototypes allow us to better analyze useful feature information from the input data. Then we learn a discriminant distance-awareness representation (DDAR) by leveraging a distinction maximization layer over optimal trainable prototypes. DDAR is simpler, more efficient, and easy to use as a pluggable layer integrating with distance-sensitive metrics. We further propose adding two constrained losses to improve the informative and discriminative properties of the latent representation. We demonstrate that DDAR addresses the feature collapse by relaxing the Lipschitz constraint hindering the practicality of DUM architectures. Through several experiments on toy examples, image classification, and text OOD detection, DDAR shows superior performance over the state-of-the-art uncertainty estimation baseline methods, specifically single-forward pass methods. Compared with the ensemble-based methods, DDAR is also competitive but more computationally efficient.

\section{Background}
% In this section, we provide a brief review of prototype learning in deep neural networks (DNNs) and deterministic uncertainty estimation methods (DUMs), which both are the foundation of our proposed methodology. 
\paragraph{Prototype Learning.}
% \subsection{Prototype learning}
% \noindent {\bf Prototype learning}. 
Prototype learning \citep{wen2016discriminative, li2021adaptive, gao2021contrastive} has been applied to feature extraction to build more discriminative features by compacting intra-class features and dispersing the inter-class ones. Specifically, few-shot learning studies are based on the prototypical networks \citep{snell2017prototypical} for their simplicity and competitive performance. Given a small support set of $N$ labeled examples $\mathcal{S}= \{(\mathbf x_1, y_1),...,(\mathbf x_N, y_N)\}$ where $\mathbf{x}_i \in \mathbb{R}^{D}$ is the feature vector nd $y_i$ is the label. Prototypical networks compute a {\em prototype}, $\mathbf c_{k}$ of each class through an embedding function $f_{\theta}:\mathbb R^{D} \rightarrow \mathbb R^{M}$. Each prototype is the mean vector of the embedded support points belonging to its class $k$:
\begin{equation}
    \mathbf c_k = \frac{1}{|\mathcal{S}_k|} \sum_{(\mathbf x_i, y_i) \in \mathcal{S}_k} f_{\theta}(\mathbf x_i), i = 1,...,K.
\end{equation}
Thus embedded new query points are classified via a softmax over distance to class prototypes:
\begin{equation}
    p_{\theta}(y = k | \mathbf x) \propto \exp(-d(f_{\theta}(\mathbf x), \mathbf c_k))
\end{equation}
where $d$ is a distance function: $\mathbb R^M \times \mathbb R^M \rightarrow [0, +\infty]$. The training is performed by minimizing the negative log-probability $\mathcal{J}(\theta) = -\log p_{\theta}(y=k|\mathbf x)$ of the true class $k$ via stochastic gradient descent (SGD). The prototype networks can be easily extended to tackle zero-shot learning where each class comes with meta-data giving a high-level description of the class rather than a small number of labeled examples. The prototype for each class can be obtained by learning an embedding of the meta-data into a shared space. 

\paragraph{Deterministic Uncertainty Estimation.}
% \subsection{Deterministic uncertainty estimation}
DUQ \citep{van2020uncertainty} builds on the RBF function which requires the preservation of input distances in the output space which is achieved using the gradient penalty. Compared with approximated GPs used in DUE \citep{van2021feature} and SNGP \citep{liu2020simple} which rely on Laplace approximation with random Fourier feature and inducing point approximation, we prefer the simpler and more efficient RBF function as the distance metric for estimating uncertainty, which is defined as the distance between the model output and the class centroids: 
% This distance is computed using the RBF kernel:  
\begin{equation}
    \mathcal{K}_c(f_{\theta}, \psi_c) = \exp \left[ -\frac{\| \mathbf{W}_c f_{\theta}(\mathbf x) - \psi_c \|_2^2 }{n \cdot 2\sigma^2}\right], \label{eq:rbf}
\end{equation}
where $f_{\theta}$ is the feature extractor parametrized by $\theta$, $\psi_c$ is the centroid for class $c$, $\mathbf{W}_c$ is a weight matrix with a length scale parameter $\sigma$, $n$ is the centroid size, and $\Psi = \{\psi_1, ...\psi_c \}$ is the class centroids. The loss function $\mathcal{L}_{RBF}$ is defined by the sum of binary cross entropy between a one-hot binary encoding of the label $y_c$ and each class kernel value $\mathcal{K}_{c}$: 
% \begin{equation}
%     \argmax_c \mathcal{K}_c(f_{\theta}(\mathbf x), \psi_c)
% \end{equation}
\begin{equation}
    \mathcal{L}_{RBF} = -\sum_{c}y_c \log(\mathcal{K}_c) + (1-y_c)\log(1-\mathcal{K}_c). \label{eq:due_loss}
\end{equation}
The training is performed by stochastic gradient descent on $\theta$ and $\mathcal{W} = \{\mathbf W_1,...,\mathbf W_c \}$. However, the loss in Eq.~\eqref{eq:due_loss} is prone to feature collapse without further regularization of DNN. Gradient penalty \citep{van2020uncertainty} can address this issue but leads to large computational overhead as it requires differentiation of the gradients of the input with respect to the DNN parameters.

\section{Methodology}
\label{sec:method}
In this section, we aim to address the three core questions by proposing a new DUM approach, based on a discriminative distance-aware representation that improves both scalability and flexibility. This is achieved by following the principle of DUMs of learning a sensitive and smooth representation but not by enforcing directly the Lipschitz constraint. 

Specifically, we start with a theoretical analysis of feature collapse and understand the essential property of the Lipschitz function. We then propose learning optimal prototypes to better capture the distance-aware property and to improve the discriminative property with the help of the distinction maximization layer. Finally, we carefully design the discriminant loss with regularization to constrain the hidden representations to mimic distances from the input space. Our DDAR method is lighter, faster and only needs a single forward pass, while it can be used as a pluggable learning layer on any top of the feature extractor, which is architecture-agnostic. 

\subsection{Feature Collapse Issue}
% \subsection{Theoretical analysis of feature collapse}
Most DUMs address the feature collapse issue through regularization methods for constraining the hidden representations to mimic distances from the input space. This is typically achieved by enforcing constraints over the Lipschitz constant of the DNN \citep{van2020uncertainty,liu2020simple}. Specifically, given the feature extractor $f_{\theta} (\mathbf x)$, the bi-Lipschitz condition implies that for any pair of inputs $\mathbf x_1$ and $\mathbf x_2$:
\begin{equation}
    L_1 \| \mathbf x_1 - \mathbf x_2 \| \le \| {f}_{\theta}(\mathbf x_1) - {f}_{\theta}(\mathbf x_2) \| \le L_2 \| \mathbf x_1 - \mathbf x_2 \| .
\end{equation}
where $L_1$ and $L_2$ are positive and bounded Lipschitz constants $0 < L_1 < 1 < L_2$. The lower Lipschitz bound $L_1$ deals with the sensitivity to preserve distances in the latent space thus avoiding feature collapse. The upper Lipschitz bound $L_2$ enforces the smoothness and robustness of a DNN by preventing over-sensitivity to perturbations in the input space of $\mathbf x$.

Typically, DUM approaches aim for bi-Lipschitz DNNs with small Lipschitz constants but this is sub-optimal based on the concentration theory \citep{boucheron2013concentration}. 

\noindent {\bf Theorem 1}. {\em Assume $\mathbf x$ is a set of random vectors, drawn from a Gaussian distribution $\mathcal{N}(0, \sigma^2 \mathbf I_d)$ and let $f: \mathbb{R}^d \rightarrow \mathbb{R}$ be a Lipschitz function with Lipschitz constant $\tau$, then we have 
\begin{equation}
    p(|f(\mathbf x) - \mathbb{E}(f(\mathbf x))| > s) \le 2\exp(-\frac{s^2}{2\tau^2\sigma^2})
\end{equation}
for all $s>0$. That means the smaller the Lipschitz constant $\tau$ is, the more the concentration of the samples around their mean increases, which results in increased feature collapse.}

This motivates us to develop a new DUM strategy that does not rely on the network to comply with the Lipschitz constraint. In the meantime, the desired Lipschitz function will separate the dissimilar samples far from each other while retaining the similar samples as closely as possible. 

\subsection{Distance-aware Latent Representation}
The DUM foundation is built upon the RBF function (or other kernel functions) which requires distance preserving \citep{van2020uncertainty}. Instead of focusing on preserving distance in the input space, we aim to deal with the distances in the latent space, named {\em distance-aware latent representation}. This is achieved by imposing a set of prototypes over the latent representations. Inspired by the prototype learning \citep{snell2017prototypical}, we leverage these prototypes to better analyze features from the new queries (samples) in light of the knowledge acquired by the DNN from the training data. Unlike the prototypical networks that use fixed prototypes (mean vector of the embedded support points), we propose to learn the optimal prototypes (as learnable parameters) for improving distance awareness in the latent space. 

Beyond the distance-aware property, the discriminative property is also critical to latent representation. The center loss \citep{wen2016discriminative} used in prototype learning can help DNN to build more discriminative features by compacting intra-class features and dispersing the inter-class ones.  In this work, we propose to use a distinction maximization (DM) layer \citep{macedo2022distinction} as a hidden layer over latent representation. This way enables us to preserve the discriminative properties of the latent representations compared to placing DM as the last layer. Compared with the softmax layer, the DM layer shows competitive performance in classification accuracy, uncertainty estimation, and OOD detection,  while maintaining deterministic neural network inference efficiency \citep{macedo2021entropic, macedo2022distinction}. 

Let's define $\mathbf z \in \mathbb{R}^n$ to be the latent representation of $\mathbf x$ given feature extractor $f_{\theta}$, i.e., $\mathbf z = f_{\theta}(\mathbf x)$, which is the input to the DM layer. Given a set of prototypes, $\mathcal{P} = \{ \mathbf p_1,...,\mathbf p_m \}$ of $m$ vectors that are trainable, we define a distinction maximization (DM) layer using cosine distance 
\begin{equation}
    \mathcal{D}_p(\mathbf z, \mathbf p_i) = \left[\frac{<\mathbf z, \mathbf{p}_1 >}{ \|\mathbf z \|_2 \| \mathbf p_1 \|_2}, \cdots, \frac{<\mathbf z, \mathbf{p}_m >}{ \|\mathbf z \|_2 \| \mathbf p_m \|_2} \right]^{\top}. \label{eq:dm}
    % \mathcal{D}_p(f_{\theta}(\mathbf x)) = \left[S_c(f_{\theta(\mathbf x)}, \mathbf p_1), \cdots, S_c(f_{\theta}(\mathbf x), \mathbf p_m) \right]^{\top}
\end{equation}
The vectors $\mathbf p_i$ in Eq.~\eqref{eq:dm} can help in better placing an input sample in the learned latent space using these prototypes as references.  Also, it is flexible to assign an arbitrarily large number of prototypes such that a richer latent mapping is defined by a finer convergence of the latent space. 

Then we apply the distinction maximization to the feature representation and subsequently use the exponential function as the activation function. This way aims to sharpen similarity values and thus facilitates the alignment of the data embedding to the prototypes. Thus the update latent representation $\tilde{f}(\theta)$ tends to be more distinctive
\begin{equation}
    \tilde{f}_{\theta} (\mathbf x) = \exp(-\mathcal{D}_{p}(f_{\theta}(\mathbf x))) = \exp(-\mathcal{D}_{p}(\mathbf z)).  \label{eq:ddar2}
\end{equation}
Note that the vector weights $\mathbf p_m$ are optimized jointly with $\theta$ and $\mathbf W_c$ in Eq.~\eqref{eq:rbf} and $\mathbf p_m$ can also work as indicators for better analyzing informative patterns in the discriminant latent representation such that DDAR is distance preserving that satisfies the bi-Lipschitz function property.  

\noindent {\bf Proposition 1}. {\em Consider a hidden mapping $f$: $\mathcal{X} \rightarrow \mathcal{F}$, the discriminant latent representation $\tilde{f}_{\theta} (\mathbf x)$ is a Lipschitz function, which satisfies 
\begin{equation}
\| \exp(-\mathcal{D}_{p}(\mathbf z_1)) - \exp(-\mathcal{D}_{p}(\mathbf z_2)) \| \le \tau \| \mathbf z_1 - \mathbf z_2 \| .
\end{equation}
Then $\tilde{f}$ is distance preserving with $\tau \in \mathbb{R}^{+}$.}
% The proof is in Appendix. 

\subsection{Discriminant Loss with Regularization Constraints}
To better address feature collapse, we improve the discrimination of latent space by adding constraints to the binary cross entropy loss in Eq.~\eqref{eq:due_loss}. To fully leverage the benefits of learnable prototypes, we propose two measures to (1) avoid the collapse of all prototypes into a single prototype. (2) not rely only on a single prototype. Thus we first add a constraint to force the prototypes to be dissimilar using the negative sum of cosine distance:
\begin{equation}
    \mathcal{L}_{{dissimilar}} = - \sum_{i < j} \frac{< \mathbf p_i, \mathbf p_j >}{\| \mathbf p_i \|_2 \| \mathbf p_j \|_2}, \label{eq:Ld}
    % S_c(\mathbf p_i, \mathbf p_j) \label{eq:Ld}
    % \mathcal{L}_{{Dissimilar}} = - \sum_{i < j} S_c(\mathbf p_i, \mathbf p_j) \label{eq:Ld}
\end{equation}
which avoids the collapse of all prototypes to a single prototype. Next, we constrain the latent representation $\mathcal{D}_{p}(f_{\theta}(\mathbf x))$ to not rely only on a single prototype by adding an entropy regularization loss: 
\begin{equation}
    \mathcal{L}_{entropy} = \sum_{k=1}^n \sigma(\mathcal{D}_{p}(\mathbf z))_k \cdot \log(\sigma(\mathcal{D}_{p}(\mathbf z))_k), \label{eq:Le}
\end{equation}
where $\sigma$ is the softmax layer, and the subscript index $k$ means the $k$-th coefficient of a tensor. Adding these two constraints in Eq.~\eqref{eq:Ld} and \eqref{eq:Le} to the RBF loss in Eq.~\eqref{eq:due_loss}, we can achieve more discriminative features by increasing the distance between prototypes and enlarging the dispersion of different prototype features. Therefore the total loss for training is defined by
\begin{equation}
    \mathcal{L}_{total} = \mathcal{L}_{RBF} + \lambda (\mathcal{L}_{dissimilar} +  \mathcal{L}_{entropy}), \label{eq:totalloss}
\end{equation}
where $\lambda \in [0,1]$ is the coefficient weight of the constraints. To avoid additional hyperparameters, we only use one coefficient to evaluate the effect of regularization constraints. Naively we can introduce another parameter to adjust the weight between $\mathcal{L}_{dissimilar}$ and $\mathcal{L}_{entropy}$. Note that we use the distinctive latent representation $\tilde{f}_{\theta}(\mathbf x)$ in Eq.~\eqref{eq:ddar2} to compute the RBF loss $\mathcal{L}_{RBF}$ rather than the original latent representation $f_{\theta}(\mathbf x)$ in Eq.~\eqref{eq:rbf}. We name this proposed method as {\em Discriminant Distance-Aware Representation} (DDAR), where the training procedure is shown in Algorithm \ref{algo:1}. 
% \subsection{DDAR algorithm}
\begin{algorithm}[h!]
% \footnotesize
\small
\caption{\hspace{-0.1cm} The {\color{black} DDAR} algorithm}
\begin{algorithmic}[1] \label{algo:1}
\STATE{\bf Requirements}: \\
- Feature extractor $f_{\theta}: x \rightarrow \mathbb{R}_d$ with feature space dimensionality $d$ and deep neural network parameters $\theta$ \\
- Hyperparameters: number of prototypes $m$, loss weight $\lambda$, length scale $\sigma$ in RBF, learning rate $\eta$, batch size $b$ \\ 
- Training and testing datasets: in-distribution data $\mathbf x_{\textup{in}}$ (e.g., CIFAR-10/100) and OOD data $\mathbf x_{\textup{ood}}$ (e.g., SVHN)
\vspace{-5pt}
\STATE{\bf Initialize:} Prototype parameters $\{\mathbf p_1,...,\mathbf p_m \}$ of $m$ vectors, $\mathbf p_i \in \mathbb{R}$ and RBF kernel weight matrix $\mathbf W_c$ (size $n \times d$)
\FOR{${\textup{train step}=1}$ \TO ${\textup{max step}}$ }
\STATE Extract the feature embedding $f_{\theta}(\mathbf x)$ 
\STATE Compute the discriminant embedding $\tilde{f}_{\theta} (\mathbf x) = \exp(-\mathcal{D}_{p}(f_{\theta}(\mathbf x)))$ after the DM layer 
\STATE Calculate the RBF loss $\mathcal{L}_{RBF}$ using Eq.~\eqref{eq:rbf}
\STATE Calculate the dissimilar loss $\mathcal{L}_{dissimilar}$ in Eq.~\eqref{eq:Ld}
\STATE Calculate the entropy loss $\mathcal{L}_{entropy}$ using Eq.~\eqref{eq:Le}
\STATE Combine all loss terms for total loss $\mathcal{L}_{total}$ in Eq.~\eqref{eq:totalloss}, $\mathcal{L}_{total} = \mathcal{L}_{RBF} + \lambda (\mathcal{L}_{dissimilar} +  \mathcal{L}_{entropy}),$
\STATE Update DNN parameters $\theta$, RBF kernel weight matrix $\mathbf W_c$ and  prototype parameters $\mathbf p$ via stochastic gradient descent: \\ 
$(\theta, \mathbf W_c, \mathbf p) \leftarrow (\theta, \mathbf W_c, \mathbf p) + \eta * \nabla_{\theta, \mathbf W_c, \mathbf p} \mathcal{L}_{total}$
\STATE Update centroids $\Psi = \{\psi_1,...,\psi_C\}$ using an exponential moving average of the feature vectors belonging to that class
\ENDFOR
\end{algorithmic}
\end{algorithm}

% \vspace{-5pt}
\section{Experiments}
\label{sec:experiments}
% \vspace{-3pt}
We show the performance of DDAR in two dimensions, with the two-moon dataset, and show the effect of discriminant distance-aware property on addressing feature collapse issues. We further test the OOD detection performance on CIFAR-10/100 vs SVHN datasets compared with multiple SOTA baselines. To verify the DDAR capability on data modalities beyond images, we also evaluate the DDAR method on practical language understanding tasks using the CLINC benchmark dataset \citep{larson2019evaluation}.  We run all baseline methods in similar settings using publicly available codes and hyperparameters for related methods. Some of the results are reported by the literature such that we can directly compare them. 
\begin{figure*}[h!]
    \centering
    \includegraphics[width=0.99\textwidth]{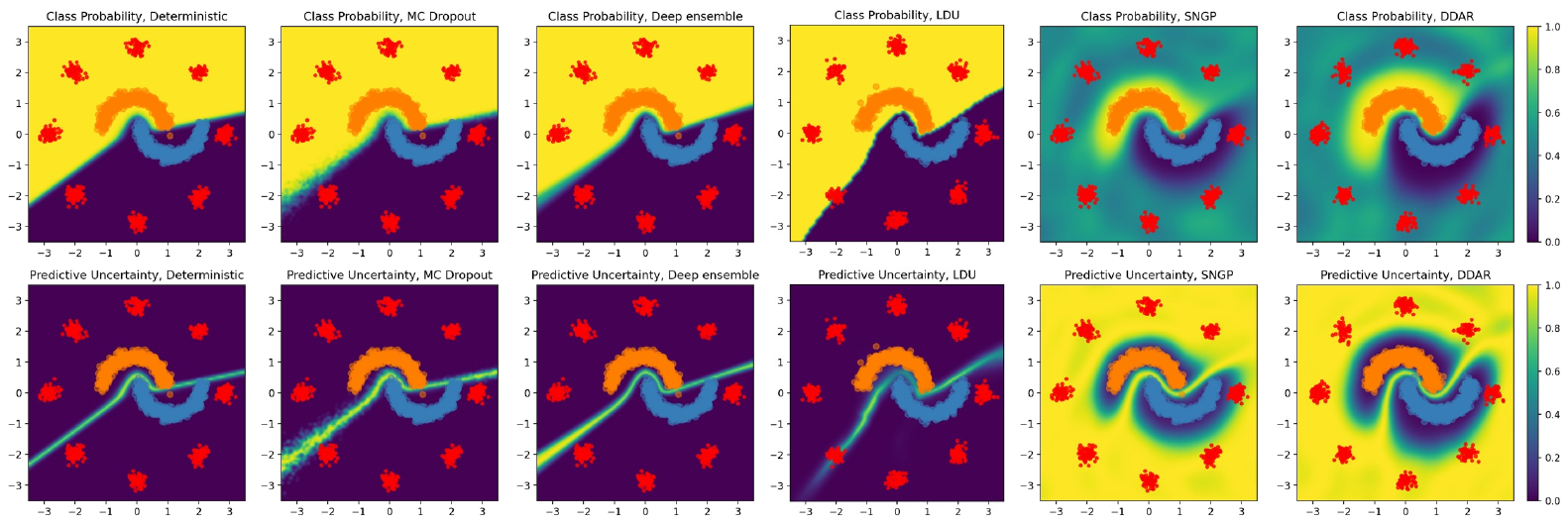}
    % \vspace{-0.2cm}
    \caption{Uncertainty results of different baseline methods on the two moon 2D classification benchmarks. Yellow indicates uncertainty, while dark blue indicates certainty. The first row is the decision boundary (class probability) and the second row is the predictive uncertainty. }
    \label{fig:toy}
\end{figure*}

\subsection{Toy Example: Two Moons}
To illustrate the DDAR method, we first consider the two moons benchmark where the dataset consists of two moon-shaped distributions separable by a nonlinear decision boundary. We use the scikit-learn implementation to draw 500 samples from each in-domain class (blue and orange dots). We use a deep feature extractor, ResFFN-12-128, which consists of 12 residual layers with 128 hidden neurons used by \citep{liu2020simple}. The embedding size is 128, the dropout rate is 0.01. We use 64 prototypes for this case and train this task using Adam optimizer with a learning rate of 0.01, batch size of 64 and set the length scale $\sigma$ of 0.3 in Eq.~\eqref{eq:rbf}, $\lambda$ of 0.1 in Eq.~\eqref{eq:totalloss}. 

Fig.~\ref{fig:toy} shows the results of decision boundary and predictive uncertainty. DDAR performs the expected behavior for high-quality predictive uncertainty: correctly classifiers the samples with high confidence (low uncertainty) in the region supported by the training data (dark blue color), and shows less confidence (high uncertainty) when samples are far away from the training data (yellow color), i.e., distance-awareness property. SNGP \citep{liu2020simple} is also able to maintain distance-awareness property via spectral normalization and shows a similar uncertainty surface. However, the other baseline methods, e.g., Deterministic (softmax) NN, MC Dropout \citep{gal2016dropout}, and Deep Ensemble \citep{lakshminarayanan2017simple}, quantify their predictive uncertainty based on the distance from the decision boundaries so only assign uncertainty along the decision boundary but certain elsewhere, which is not distance aware. They are overconfident since they assign high certainty to OOD samples even if they are far from the data. LDU \citep{macedo2022distinction} shows a similar overconfident result without leveraging OOD samples into training. 

\begin{figure}[h!]
    \centering
    \includegraphics[width=0.48\textwidth]{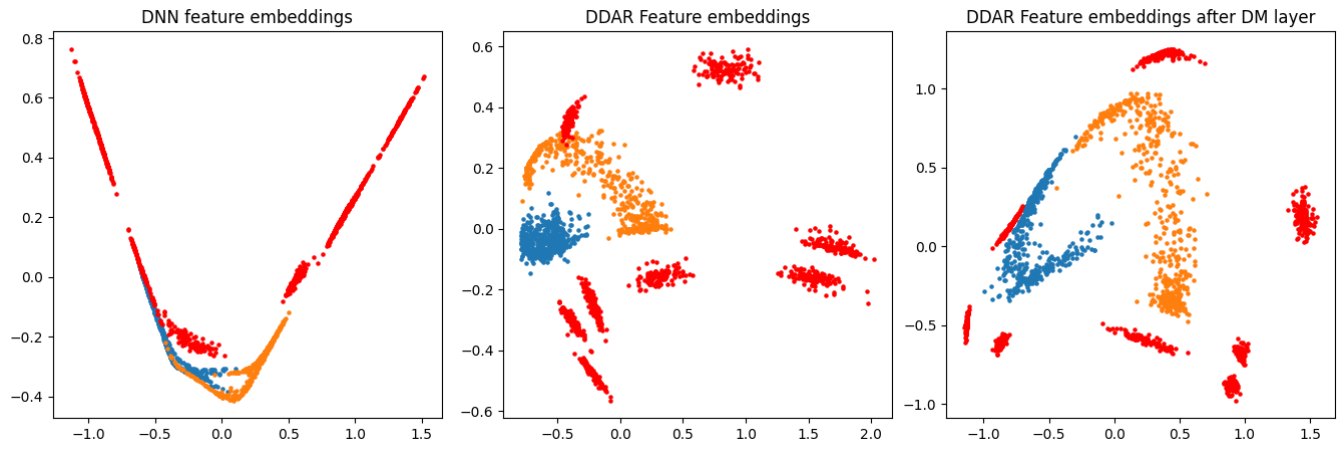}
    % \vspace{-0.4cm}
    \caption{Addressing feature collapse. 2D project of feature embedding of regular DNN (left), DDAR embedding (middle), and DDAR embedding after the DM layer, trained on the two moons dataset. }
    \label{fig:projection}
\end{figure}

\noindent {\bf Feature Collapse}. Fig.~\ref{fig:projection} shows the PCA projection of the feature embedding of two moons dataset for feature collapse illustration. The objective for the regular DNN model introduces a large amount of distortion of the space, collapsing the two-class samples and OOD samples to a single line, making it almost impossible to use distance-awareness metric on these features. Specifically, the OOD samples move from a separated area in the input space on top of class data in the feature space, which fails in OOD detection tasks and results in unreliable predictive uncertainty estimation. In contrast, the feature embedding learned by the DDAR method allows a better disentangling of the latent space without overlapping the two classes. Furthermore, the learned embedding after the DM layer accurately maintains the relative distances of the two classes and OOD data. 
\begin{table*}[!h]
\footnotesize
% \scriptsize
\centering
\caption{OOD detection results: CIFAR-10 vs SVHN OOD.}
\label{tab:cifar10}
\vspace{-0.3cm}
\begin{tabular}{l|ccc}
\toprule
Method        & Accuracy ($\uparrow$)    & ECE ($\downarrow$)           & AUROC ($\uparrow$)  \\ \hline 
DNN (Softmax)  & 94.3 $\pm$ 0.11 & 0.024 $\pm$ 0.003 & 0.928 $\pm$ 0.005 \\
MC Dropout \citep{gal2016dropout}    & 95.7 $\pm$ 0.13 & 0.013 $\pm$ 0.002 & 0.934 $\pm$ 0.004   \\
Deep Ensemble \citep{lakshminarayanan2017simple} & {\bf 96.4 $\pm$ 0.09} & {\bf 0.011 $\pm$ 0.001} & {\bf 0.947 $\pm$ 0.002}      \\
DUQ \citep{van2020uncertainty}          & 95.8 $\pm$ 0.12 & 0.027 $\pm$ 0.001 & 0.939 $\pm$ 0.007      \\
SNGP \citep{liu2020simple}         & 95.7 $\pm$ 0.14 & 0.017 $\pm$ 0.003 & 0.940 $\pm$ 0.004      \\
LDU \citep{macedo2022distinction} & 95.5 $\pm$ 0.17 & { 0.021 $\pm$ 0.002} & 0.945 $\pm$ 0.004 \\ 
DDU \citep{mukhoti2021deep} & 95.1 $\pm$ 0.12 & {\bf 0.014 $\pm$ 0.002} & 0.936 $\pm$ 0.003 \\ 
MIR \citep{postels2020quantifying}  & 94.9 $\pm$ 0.15 & 0.021 $\pm$ 0.003 & 0.912 $\pm$ 0.005 \\ 
\hline
DDAR with 256 prototypes         & {\bf 96.0 $\pm$ 0.12}           & 0.015 $\pm$ 0.002             & {\bf 0.949 $\pm$ 0.003}              \\
DDAR with 64 prototypes         & 95.8 $\pm$ 0.13          & 0.016 $\pm$ 0.002             & {\bf 0.947 $\pm$ 0.003}               \\ \toprule
\end{tabular}
\end{table*}

\begin{table*}[!h]
\footnotesize
% \scriptsize
\centering
\caption{OOD detection results: CIFAR-100 vs SVHN OOD.}
\label{tab:cifar100}
\vspace{-0.3cm}
\begin{tabular}{l|ccc}
\toprule
Method        & Accuracy ($\uparrow$)    & ECE ($\downarrow$)           & AUROC ($\uparrow$)  \\ \hline 
DNN (Softmax)  & 80.4 $\pm$ 0.11 & 0.082 $\pm$ 0.002 & 0.763 $\pm$ 0.011 \\
MC Dropout \citep{gal2016dropout}    & 80.2 $\pm$ 0.22 & {\bf 0.031 $\pm$ 0.002} & 0.800 $\pm$ 0.014   \\
Deep Ensemble \citep{lakshminarayanan2017simple} & {\bf 82.5 $\pm$ 0.19} & 0.041 $\pm$ 0.002 & {\bf 0.832 $\pm$ 0.007}      \\
DUQ \citep{van2020uncertainty}          & 79.7 $\pm$ 0.20 & 0.112 $\pm$ 0.002 & 0.777 $\pm$ 0.026      \\
SNGP \citep{liu2020simple}         & {\bf 82.5 $\pm$ 0.16} & {\bf 0.030 $\pm$ 0.004} & 0.821 $\pm$ 0.019      \\
LDU \citep{macedo2022distinction} & 81.3 $\pm$ 0.15 & {0.052 $\pm$ 0.003} & 0.822 $\pm$ 0.003 \\ 
DDU \citep{mukhoti2021deep}  & 81.6 $\pm$ 0.14 & 0.029 $\pm$ 0.003 & 0.826 $\pm$ 0.009 \\ 
MIR \citep{postels2020quantifying} & 80.9 $\pm$ 0.18 & 0.037 $\pm$ 0.002 & 0.788 $\pm$ 0.011 \\ 
\hline
DDAR with 256 prototypes           & {\bf 82.5 $\pm$ 0.17}           & 0.032 $\pm$ 0.002             & {\bf 0.829 $\pm$ 0.008}                \\
DDAR with 64 prototypes          & 82.0 $\pm$ 0.17           & 0.035 $\pm$ 0.002             & 0.826  $\pm$ 0.009              \\ \toprule
\end{tabular}
\end{table*}

\subsection{OOD Detection: CIFAR-10/100 vs SVHN}
\noindent {\bf Baseline Methods}. We compare DDAR with two baselines for uncertainty estimation - MC Dropout \citep{gal2016dropout} (with 10 dropout samples) and Deep Ensemble \citep{lakshminarayanan2017simple} (with 10 models). We also include the softmax entropy of a regular DNN as a simple baseline. We choose three deterministic uncertainty methods (DUMs) - DUQ \citep{van2020uncertainty}, SNGP \citep{liu2020simple}, and LDU \citep{macedo2022distinction} as representatives of distance-awareness (discriminative approaches) for uncertainty estimation. Also, we evaluate DDU \citep{mukhoti2021deep} and MIR \citep{postels2020quantifying} as representatives of informative representation (generative approaches \citep{winkens2020contrastive,charpentier2020posterior}), which fit a class-conditional GMM to their regularized latent representations and use the loglikelihood as an uncertainty proxy \citep{postels2021practicality}. For DDAR, we consider two cases with 256 (DDAR-256) and 64 (DDAR-64) prototypes respectively. 

\noindent  {\bf Datasets}. We train the DDAR model on CIFAR-10 and CIFAR-100 image classification tasks. Following the benchmarking setup suggested in \citep{ovadia2019can}, we evaluate the model's predictive accuracy and expected calibration error (ECE) \citep{guo2017calibration} under clean CIFAR-10/100 testing data. To evaluate the model's OOD performance, we choose the standard OOD task \citep{van2020uncertainty} using SVHN as the OOD data for a model trained on CIFAR-10/100. We normalize the OOD datasets using the in-distribution training data (CIFAR-10/100) and use the Area Under the Reciever Operator Curve (AUROC) metric to report the OOD detection (image classification) performance. 

\noindent  {\bf Model and Optimization}. Each baseline method shares the same ResFFN-12-128 architecture as used in two moons for training on CIFAR-10/100 datasets. Each method has a hyperparameter for the regularization of its latent representation, and we choose the hyperparameter so that it minimizes the validation loss. To compare the variation of each method, all results are averaged over 5 independent runs on an NVIDIA V100 GPU. We use 64 and 256 prototypes for each case and train the tasks using Adam optimizer with a learning rate of 0.01, and batch size of 128. 

\noindent {\bf Results}. To evaluate OOD detection performance, we use the standard setting based on training on CIFAR-10 and CIFAR-100 as in-distribution data and SVHN as OOD data. The performance of different baselines for CIFAR-10 and CIFAR-100 are shown in Table \ref{tab:cifar10} and \ref{tab:cifar100} (The top 2 results are highlighted in bold). Note that our DDAR method shows superior results, specifically on the AUROC and Accuracy metrics. With the increasing of prototypes, DDAR shows improved performance on OOD detection. This is because more prototypes increase the flexibility to model complex distributions of the discriminant latent representation. 
% \vspace{-0.5cm}
\begin{table}[!h]
\footnotesize
% \script
\centering
\caption{Ablation study of loss weight $\lambda$ on CIFAR-100.}
\label{tab:weight}
\vspace{-0.3cm}
\begin{tabular}{c|ccc}
\toprule
$\lambda$        & Accuracy ($\uparrow$)    & ECE ($\downarrow$)           & AUROC ($\uparrow$)  \\ \hline 
0.01 & {81.9 $\pm$ 0.18}           & {0.043 $\pm$ 0.003}             & {0.801 $\pm$ 0.012}  \\
0.05     & {82.1 $\pm$ 0.16}           & {0.037 $\pm$ 0.002}             & {0.817 $\pm$ 0.009}    \\
0.1  & {\bf 82.5 $\pm$ 0.17}           & {\bf 0.032 $\pm$ 0.002}             & {\bf 0.829 $\pm$ 0.008}      \\
0.5           & {82.4 $\pm$ 0.15}           & {0.039 $\pm$ 0.003}             & {0.824 $\pm$ 0.006}      \\
1.0          & {\bf 82.5 $\pm$ 0.17}           & {0.035 $\pm$ 0.002}             & {0.820 $\pm$ 0.007}                 \\ \toprule
\end{tabular}
\end{table}
% \vspace{-0.5cm}
\begin{table}[!h]
% \footnotesize
\scriptsize
\centering
\caption{Effect of loss given $\lambda=0.1$ on CIFAR-100.}
\label{tab:loss}
\vspace{-0.3cm}
\begin{tabular}{c|ccc}
\toprule
Loss        & Accuracy ($\uparrow$)    & ECE ($\downarrow$)           & AUROC ($\uparrow$)  \\ \hline 
$\mathcal{L}_{d}$ & {81.3 $\pm$ 0.13}           & 0.041 $\pm$ 0.003             & {0.813 $\pm$ 0.009} \\
$\mathcal{L}_{e}$    & {82.0 $\pm$ 0.19}           & 0.033 $\pm$ 0.002             & {0.822 $\pm$ 0.007}   \\
$\mathcal{L}_{d}$ \& $\mathcal{L}_{e}$  & {\bf 82.5 $\pm$ 0.17}           & \bf 0.032 $\pm$ 0.002             & {\bf 0.829 $\pm$ 0.008}   \\ \toprule
\end{tabular}
\end{table}

\begin{table*}[!h]
\footnotesize
% \scriptsize
\centering
\caption{OOD detection results for BERT on CLINC dataset, averaged over 10 random trials.}
\label{tab:nlp}
\vspace{-0.3cm}
\begin{tabular}{l|ccc}
\toprule
Method        & Accuracy ($\uparrow$)    & ECE ($\downarrow$)           & AUROC ($\uparrow$)  \\ \hline 
DNN (Softmax)  & 96.5 $\pm$ 0.11 & 0.024 $\pm$ 0.002 & 0.897 $\pm$ 0.01 \\
MC Dropout \citep{gal2016dropout}    & 96.1 $\pm$ 0.10 & {0.021 $\pm$ 0.001} & 0.938 $\pm$ 0.01   \\
Deep Ensemble \citep{lakshminarayanan2017simple} & {\bf 97.5 $\pm$ 0.03} & {\bf 0.013 $\pm$ 0.002} & {0.964 $\pm$ 0.01}      \\
DUQ \citep{van2020uncertainty}          & 96.0 $\pm$ 0.04 & 0.059 $\pm$ 0.002 & 0.917 $\pm$ 0.01      \\
SNGP \citep{liu2020simple}         & {96.6 $\pm$ 0.05} & {\bf 0.014 $\pm$ 0.005} & {\bf 0.969 $\pm$ 0.01}      \\
% LDU \citep{macedo2022distinction} & 81.3 $\pm$ 0.15 & {0.052 $\pm$ 0.003} & 0.822 $\pm$ 0.003 \\ 
% DDU \citep{mukhoti2021deep}  & 81.6 $\pm$ 0.14 & 0.029 $\pm$ 0.003 & 0.826 $\pm$ 0.009 \\ 
% MIR \citep{postels2020quantifying} & 80.9 $\pm$ 0.18 & 0.037 $\pm$ 0.002 & 0.788 $\pm$ 0.011 \\ 
\hline
DDAR with 256 prototypes           & {\bf 97.5 $\pm$ 0.04}           & {\bf 0.014 $\pm$ 0.002}             & {\bf 0.969 $\pm$ 0.01}                \\
DDAR with 64 prototypes          & 96.3 $\pm$ 0.05           & 0.017 $\pm$ 0.002             & 0.961  $\pm$ 0.02              \\ \toprule
\end{tabular}
\end{table*}

\noindent {\bf Ablation Studies}.
We further investigate the effect of constrained loss terms and the corresponding weights on the OOD detection performance. Table \ref{tab:weight} and Table \ref{tab:loss} show the ablation studies on the choice of loss weight $\lambda$ and loss terms $\mathcal{L}_{dissimilar}$ ($\mathcal{L}_d$) and $\mathcal{L}_{entropy}$  ($\mathcal{L}_e$) on CIFAR-100 task respectively. Note that the loss weight $\lambda=0.1$ is the best choice for this case and the two constrained losses contribute consistent improvements individually and together. 
% \noindent {\bf Ablation study}. 

% \vspace{-5pt}
\subsection{Conversational Language Understanding}
To demonstrate the effectiveness of the proposed distance-awareness representation on data modalities beyond images, we further evaluate the DDAR method on practical language understanding tasks where uncertainty estimation is highly critical: detecting out-of-scope dialog intent \citep{larson2019evaluation,vedula2019towards,yaghoub2020user,zheng2020out}. For a dialog system (e.g., chatbot) built for in-domain services, it is essential to understand if an input natural utterance from a user is out-of-scope or in-scope. In other words, the model should know when it abstains from or activates one of the in-domain services. To this end, this problem can be formulated as an OOD detection problem where we consider training an intent understanding model to detect in-domain services or out-of-domain services. 

We follow the problem setup \citep{liu2020simple} and use the CLINC out-of-scope intent detection benchmark dataset \citep{larson2019evaluation} which contains 150 in-domain services data with 150 training sentences in each domain, and 1500 natural out-of-domain utterances. We train a BERT model only on in-domain data and evaluate their predictive accuracy on the in-domain test data, their calibration, and OOD detection performance on the combined in-domain and out-of-domain data. The results are shown in Table \ref{tab:nlp}. In this case, we only compare the baseline methods, including DNN, MC Dropout, Deep Ensemble, DUQ, and SNGP. As shown, consistent with the previous vision experiments, our DDAR method shows competitive performance, which outperforms other single model approaches and is close to the deep ensemble in prediction accuracy and to SNGP in confidence calibration and OOD AUROC. 

% \vspace{-5pt}
\section{Related Work}
% \vspace{-3pt}
\paragraph{Deterministic Uncertainty Methods.} 
Unlike the conventional uncertainty estimation methods, including Bayesian Neural Networks (BNNs) \citep{osawa2019practical, wenzel2020good}, MC Dropout \citep{gal2016dropout}, and Deep Ensemble \citep{lakshminarayanan2017simple}, a promising line of work recently emerged for estimating uncertainties of a DNN with a single forward pass while treating its weights deterministically \citep{postels2021practicality}. By regularizing the hidden representations of a model, these methods represent an efficient and scalable solution to uncertainty estimation and to the related out-of-distribution (OOD) detection problem. In contrast to BNNs, Deterministic Uncertainty Methods (DUMs) quantify epistemic uncertainty using the distribution of latent representations \citep{alemi2018uncertainty,wu2020simple,charpentier2020posterior,mukhoti2021deterministic,charpentier2021natural} or by
replacing the final softmax layer with a distance-sensitive
function \citep{mandelbaum2017distance, van2020uncertainty,liu2020simple,van2021feature}.  Note that there is another line of work, which proposes a principled approach for variance propagation in DNNs \citep{postels2019sampling,haussmann2020sampling,loquercio2020general} but these approaches fundamentally differ from DUMs due to their probabilistic treatment of the parameters, even though they are efficient approaches and also relied on a single forward pass for uncertainty estimation. 

\paragraph{Addressing Feature Collapse.} 
The critical challenge in DUMs is how to address the feature collapse issue. Currently, there are two main paradigms - distance awareness and informative representations. The distance awareness avoids feature collapse by relating distances between latent representations to distance in the input space. The primary methods are to impose the bi-Lipschitz constraint by using a two-sided gradient penalty \citep{van2020uncertainty} or spectral normalization \citep{liu2020simple,van2021feature,mukhoti2021deep}. While distance-awareness achieves remarkable performance on OOD detection, it does not explicitly preserve sample-specific information. An alternative line of work addresses this challenge by learning informative representations \citep{alemi2018uncertainty,wu2020simple,postels2020hidden}, thus forcing discriminative models to preserve information in its hidden representations. While distance-awareness is
based on the choice of a specific distance metric tying together input and latent space, informative representations
incentivize a DNN to store more information about the input. There are several approaches in this paradigm, including contrastive learning \citep{chen2020simple,wu2020simple,winkens2020contrastive}, reconstruction regularization \citep{postels2020hidden}, entropy regularization \citep{charpentier2020posterior}, and invertible neural networks \citep{behrmann2019invertible,ardizzone2020training,nalisnick2019hybrid,ardizzone2018analyzing}. 

% \paragraph{Prototype learning.} Recently, a set of promising works, named Deterministic Uncertainty Methods (DUMs) \citep{mukhoti2021deep} emerged for estimating uncertainty with a single forward pass while treating its weights deterministically \citep{postels2021practicality}. These methods are prone to be efficient and scalable solutions to uncertainty estimation and out-of-distribution (OOD) detection problems. DUMs aim at learning informative latent representation of a model given that the distribution of latent representation should be representative of the input distribution. Then DUMs estimate uncertainty by replacing the final softmax layer with a distance-sensitive function. Specifically, DUQ \citep{van2020uncertainty} defines the uncertainty as the distance between the model output and the closest centroid and proposes a novel centroid updating step based on Radial Basis Function (RBF) networks. 
% \vspace{-5pt}
\section{Conclusions}
\label{sec:conclusions}
% \vspace{-3pt}
In this work, we develop a novel {DDAR} method for deterministic uncertainty estimation. This is achieved by learning a discriminant distance-aware representation that leverages a distinction maximization layer over a set of learnable prototypes. Compared with the baseline DUMs, our DDAR is a simple and efficient method without the feature collapse issue while the feature extractor is architecture-agnostic with higher flexibility not limited by residual neural networks. Through several experiments on synthesis data, image classification, and text OOD detection benchmarks, we show that DDAR outperforms the different SOTA baselines in terms of prediction accuracy, confidence calibration, and OOD detection performance. 

The limitation of this work is a lack of a deep theoretical understanding of feature collapse although the empirical improvements are clearly shown. We plan to dig into the theoretical propriety of feature collapse with better model interpretability and explainability. Future work could also investigate the scalability of DDAR on large-scale computer vision tasks, and large language models (LLMs).

\bibliographystyle{plainnat}
\bibliography{reference.bib}

\begin{thebibliography}{45}
\providecommand{\natexlab}[1]{#1}
\providecommand{\url}[1]{\texttt{#1}}
\expandafter\ifx\csname urlstyle\endcsname\relax
  \providecommand{\doi}[1]{doi: #1}\else
  \providecommand{\doi}{doi: \begingroup \urlstyle{rm}\Url}\fi

\bibitem[Alemi et~al.(2018)Alemi, Fischer, and Dillon]{alemi2018uncertainty}
Alexander~A Alemi, Ian Fischer, and Joshua~V Dillon.
\newblock Uncertainty in the variational information bottleneck.
\newblock \emph{arXiv preprint arXiv:1807.00906}, 2018.

\bibitem[Ardizzone et~al.(2018)Ardizzone, Kruse, Wirkert, Rahner, Pellegrini,
  Klessen, Maier-Hein, Rother, and K{\"o}the]{ardizzone2018analyzing}
Lynton Ardizzone, Jakob Kruse, Sebastian Wirkert, Daniel Rahner, Eric~W
  Pellegrini, Ralf~S Klessen, Lena Maier-Hein, Carsten Rother, and Ullrich
  K{\"o}the.
\newblock Analyzing inverse problems with invertible neural networks.
\newblock \emph{arXiv preprint arXiv:1808.04730}, 2018.

\bibitem[Ardizzone et~al.(2020)Ardizzone, Mackowiak, Rother, and
  K{\"o}the]{ardizzone2020training}
Lynton Ardizzone, Radek Mackowiak, Carsten Rother, and Ullrich K{\"o}the.
\newblock Training normalizing flows with the information bottleneck for
  competitive generative classification.
\newblock \emph{Advances in Neural Information Processing Systems},
  33:\penalty0 7828--7840, 2020.

\bibitem[Behrmann et~al.(2019)Behrmann, Grathwohl, Chen, Duvenaud, and
  Jacobsen]{behrmann2019invertible}
Jens Behrmann, Will Grathwohl, Ricky~TQ Chen, David Duvenaud, and
  J{\"o}rn-Henrik Jacobsen.
\newblock Invertible residual networks.
\newblock In \emph{International Conference on Machine Learning}, pages
  573--582. PMLR, 2019.

\bibitem[Boucheron et~al.(2013)Boucheron, Lugosi, and
  Massart]{boucheron2013concentration}
St{\'e}phane Boucheron, G{\'a}bor Lugosi, and Pascal Massart.
\newblock \emph{Concentration inequalities: A nonasymptotic theory of
  independence}.
\newblock Oxford university press, 2013.

\bibitem[Charpentier et~al.(2020)Charpentier, Z{\"u}gner, and
  G{\"u}nnemann]{charpentier2020posterior}
Bertrand Charpentier, Daniel Z{\"u}gner, and Stephan G{\"u}nnemann.
\newblock Posterior network: Uncertainty estimation without ood samples via
  density-based pseudo-counts.
\newblock \emph{Advances in Neural Information Processing Systems},
  33:\penalty0 1356--1367, 2020.

\bibitem[Charpentier et~al.(2021)Charpentier, Borchert, Z{\"u}gner, Geisler,
  and G{\"u}nnemann]{charpentier2021natural}
Bertrand Charpentier, Oliver Borchert, Daniel Z{\"u}gner, Simon Geisler, and
  Stephan G{\"u}nnemann.
\newblock Natural posterior network: Deep bayesian uncertainty for exponential
  family distributions.
\newblock \emph{arXiv preprint arXiv:2105.04471}, 2021.

\bibitem[Chen et~al.(2020)Chen, Kornblith, Norouzi, and Hinton]{chen2020simple}
Ting Chen, Simon Kornblith, Mohammad Norouzi, and Geoffrey Hinton.
\newblock A simple framework for contrastive learning of visual
  representations.
\newblock In \emph{International conference on machine learning}, pages
  1597--1607. PMLR, 2020.

\bibitem[Franchi et~al.(2022)Franchi, Yu, Bursuc, Aldea, Dubuisson, and
  Filliat]{franchi2022latent}
Gianni Franchi, Xuanlong Yu, Andrei Bursuc, Emanuel Aldea, Severine Dubuisson,
  and David Filliat.
\newblock Latent discriminant deterministic uncertainty.
\newblock \emph{arXiv preprint arXiv:2207.10130}, 2022.

\bibitem[Gal and Ghahramani(2016)]{gal2016dropout}
Yarin Gal and Zoubin Ghahramani.
\newblock Dropout as a bayesian approximation: Representing model uncertainty
  in deep learning.
\newblock In \emph{international conference on machine learning}, pages
  1050--1059. PMLR, 2016.

\bibitem[Gao et~al.(2021)Gao, Fei, Liu, Lu, and Xiang]{gao2021contrastive}
Yizhao Gao, Nanyi Fei, Guangzhen Liu, Zhiwu Lu, and Tao Xiang.
\newblock Contrastive prototype learning with augmented embeddings for few-shot
  learning.
\newblock In \emph{Uncertainty in Artificial Intelligence}, pages 140--150.
  PMLR, 2021.

\bibitem[Gawlikowski et~al.(2023)Gawlikowski, Tassi, Ali, Lee, Humt, Feng,
  Kruspe, Triebel, Jung, Roscher, et~al.]{gawlikowski2023survey}
Jakob Gawlikowski, Cedrique Rovile~Njieutcheu Tassi, Mohsin Ali, Jongseok Lee,
  Matthias Humt, Jianxiang Feng, Anna Kruspe, Rudolph Triebel, Peter Jung,
  Ribana Roscher, et~al.
\newblock A survey of uncertainty in deep neural networks.
\newblock \emph{Artificial Intelligence Review}, 56\penalty0 (Suppl
  1):\penalty0 1513--1589, 2023.

\bibitem[Guo et~al.(2017)Guo, Pleiss, Sun, and Weinberger]{guo2017calibration}
Chuan Guo, Geoff Pleiss, Yu~Sun, and Kilian~Q Weinberger.
\newblock On calibration of modern neural networks.
\newblock In \emph{International conference on machine learning}, pages
  1321--1330. PMLR, 2017.

\bibitem[Hau{\ss}mann et~al.(2020)Hau{\ss}mann, Hamprecht, and
  Kandemir]{haussmann2020sampling}
Manuel Hau{\ss}mann, Fred~A Hamprecht, and Melih Kandemir.
\newblock Sampling-free variational inference of bayesian neural networks by
  variance backpropagation.
\newblock In \emph{Uncertainty in Artificial Intelligence}, pages 563--573.
  PMLR, 2020.

\bibitem[Lakshminarayanan et~al.(2017)Lakshminarayanan, Pritzel, and
  Blundell]{lakshminarayanan2017simple}
Balaji Lakshminarayanan, Alexander Pritzel, and Charles Blundell.
\newblock Simple and scalable predictive uncertainty estimation using deep
  ensembles.
\newblock \emph{Advances in neural information processing systems}, 30, 2017.

\bibitem[Larson et~al.(2019)Larson, Mahendran, Peper, Clarke, Lee, Hill,
  Kummerfeld, Leach, Laurenzano, Tang, et~al.]{larson2019evaluation}
Stefan Larson, Anish Mahendran, Joseph~J Peper, Christopher Clarke, Andrew Lee,
  Parker Hill, Jonathan~K Kummerfeld, Kevin Leach, Michael~A Laurenzano,
  Lingjia Tang, et~al.
\newblock An evaluation dataset for intent classification and out-of-scope
  prediction.
\newblock \emph{arXiv preprint arXiv:1909.02027}, 2019.

\bibitem[Li et~al.(2021)Li, Jampani, Sevilla-Lara, Sun, Kim, and
  Kim]{li2021adaptive}
Gen Li, Varun Jampani, Laura Sevilla-Lara, Deqing Sun, Jonghyun Kim, and
  Joongkyu Kim.
\newblock Adaptive prototype learning and allocation for few-shot segmentation.
\newblock In \emph{Proceedings of the IEEE/CVF Conference on Computer Vision
  and Pattern Recognition}, pages 8334--8343, 2021.

\bibitem[Liu et~al.(2020)Liu, Lin, Padhy, Tran, Bedrax~Weiss, and
  Lakshminarayanan]{liu2020simple}
Jeremiah Liu, Zi~Lin, Shreyas Padhy, Dustin Tran, Tania Bedrax~Weiss, and
  Balaji Lakshminarayanan.
\newblock Simple and principled uncertainty estimation with deterministic deep
  learning via distance awareness.
\newblock \emph{Advances in Neural Information Processing Systems},
  33:\penalty0 7498--7512, 2020.

\bibitem[Loquercio et~al.(2020)Loquercio, Segu, and
  Scaramuzza]{loquercio2020general}
Antonio Loquercio, Mattia Segu, and Davide Scaramuzza.
\newblock A general framework for uncertainty estimation in deep learning.
\newblock \emph{IEEE Robotics and Automation Letters}, 5\penalty0 (2):\penalty0
  3153--3160, 2020.

\bibitem[Mac{\^e}do et~al.(2021)Mac{\^e}do, Ren, Zanchettin, Oliveira, and
  Ludermir]{macedo2021entropic}
David Mac{\^e}do, Tsang~Ing Ren, Cleber Zanchettin, Adriano~LI Oliveira, and
  Teresa Ludermir.
\newblock Entropic out-of-distribution detection.
\newblock In \emph{2021 International Joint Conference on Neural Networks
  (IJCNN)}, pages 1--8. IEEE, 2021.

\bibitem[Mac{\^e}do et~al.(2022)Mac{\^e}do, Zanchettin, and
  Ludermir]{macedo2022distinction}
David Mac{\^e}do, Cleber Zanchettin, and Teresa Ludermir.
\newblock Distinction maximization loss: Efficiently improving classification
  accuracy, uncertainty estimation, and out-of-distribution detection simply
  replacing the loss and calibrating.
\newblock \emph{arXiv preprint arXiv:2205.05874}, 2022.

\bibitem[Mandelbaum and Weinshall(2017)]{mandelbaum2017distance}
Amit Mandelbaum and Daphna Weinshall.
\newblock Distance-based confidence score for neural network classifiers.
\newblock \emph{arXiv preprint arXiv:1709.09844}, 2017.

\bibitem[Miyato et~al.(2018)Miyato, Kataoka, Koyama, and
  Yoshida]{miyato2018spectral}
Takeru Miyato, Toshiki Kataoka, Masanori Koyama, and Yuichi Yoshida.
\newblock Spectral normalization for generative adversarial networks.
\newblock In \emph{International Conference on Learning Representations}, 2018.

\bibitem[Mukhoti et~al.(2021{\natexlab{a}})Mukhoti, Kirsch, van Amersfoort,
  Torr, and Gal]{mukhoti2021deep}
Jishnu Mukhoti, Andreas Kirsch, Joost van Amersfoort, Philip~HS Torr, and Yarin
  Gal.
\newblock Deep deterministic uncertainty: A simple baseline.
\newblock \emph{arXiv e-prints}, pages arXiv--2102, 2021{\natexlab{a}}.

\bibitem[Mukhoti et~al.(2021{\natexlab{b}})Mukhoti, Kirsch, van Amersfoort,
  Torr, and Gal]{mukhoti2021deterministic}
Jishnu Mukhoti, Andreas Kirsch, Joost van Amersfoort, Philip~HS Torr, and Yarin
  Gal.
\newblock Deterministic neural networks with appropriate inductive biases
  capture epistemic and aleatoric uncertainty.
\newblock \emph{arXiv preprint arXiv:2102.11582}, 2021{\natexlab{b}}.

\bibitem[Nalisnick et~al.(2019)Nalisnick, Matsukawa, Teh, Gorur, and
  Lakshminarayanan]{nalisnick2019hybrid}
Eric Nalisnick, Akihiro Matsukawa, Yee~Whye Teh, Dilan Gorur, and Balaji
  Lakshminarayanan.
\newblock Hybrid models with deep and invertible features.
\newblock In \emph{International Conference on Machine Learning}, pages
  4723--4732. PMLR, 2019.

\bibitem[Osawa et~al.(2019)Osawa, Swaroop, Khan, Jain, Eschenhagen, Turner, and
  Yokota]{osawa2019practical}
Kazuki Osawa, Siddharth Swaroop, Mohammad Emtiyaz~E Khan, Anirudh Jain, Runa
  Eschenhagen, Richard~E Turner, and Rio Yokota.
\newblock Practical deep learning with bayesian principles.
\newblock \emph{Advances in neural information processing systems}, 32, 2019.

\bibitem[Ovadia et~al.(2019)Ovadia, Fertig, Ren, Nado, Sculley, Nowozin,
  Dillon, Lakshminarayanan, and Snoek]{ovadia2019can}
Yaniv Ovadia, Emily Fertig, Jie Ren, Zachary Nado, David Sculley, Sebastian
  Nowozin, Joshua Dillon, Balaji Lakshminarayanan, and Jasper Snoek.
\newblock Can you trust your model's uncertainty? evaluating predictive
  uncertainty under dataset shift.
\newblock \emph{Advances in neural information processing systems}, 32, 2019.

\bibitem[Postels et~al.()Postels, Blum, Cadena, Siegwart, Van~Gool, and
  Tombari]{postels2020quantifying}
Janis Postels, Hermann Blum, Cesar Cadena, Roland Siegwart, Luc Van~Gool, and
  Federico Tombari.
\newblock Quantifying aleatoric and epistemic uncertainty using density
  estimation in latent space.

\bibitem[Postels et~al.(2019)Postels, Ferroni, Coskun, Navab, and
  Tombari]{postels2019sampling}
Janis Postels, Francesco Ferroni, Huseyin Coskun, Nassir Navab, and Federico
  Tombari.
\newblock Sampling-free epistemic uncertainty estimation using approximated
  variance propagation.
\newblock In \emph{Proceedings of the IEEE/CVF International Conference on
  Computer Vision}, pages 2931--2940, 2019.

\bibitem[Postels et~al.(2020)Postels, Blum, Str{\"u}mpler, Cadena, Siegwart,
  Van~Gool, and Tombari]{postels2020hidden}
Janis Postels, Hermann Blum, Yannick Str{\"u}mpler, Cesar Cadena, Roland
  Siegwart, Luc Van~Gool, and Federico Tombari.
\newblock The hidden uncertainty in a neural networks activations.
\newblock \emph{arXiv preprint arXiv:2012.03082}, 2020.

\bibitem[Postels et~al.(2021)Postels, Segu, Sun, Van~Gool, Yu, and
  Tombari]{postels2021practicality}
Janis Postels, Mattia Segu, Tao Sun, Luc Van~Gool, Fisher Yu, and Federico
  Tombari.
\newblock On the practicality of deterministic epistemic uncertainty.
\newblock \emph{arXiv preprint arXiv:2107.00649}, 2021.

\bibitem[Snell et~al.(2017)Snell, Swersky, and Zemel]{snell2017prototypical}
Jake Snell, Kevin Swersky, and Richard Zemel.
\newblock Prototypical networks for few-shot learning.
\newblock \emph{Advances in neural information processing systems}, 30, 2017.

\bibitem[Tran et~al.(2022)Tran, Liu, Dusenberry, Phan, Collier, Ren, Han, Wang,
  Mariet, Hu, et~al.]{tran2022plex}
Dustin Tran, Jeremiah Liu, Michael~W Dusenberry, Du~Phan, Mark Collier, Jie
  Ren, Kehang Han, Zi~Wang, Zelda Mariet, Huiyi Hu, et~al.
\newblock Plex: Towards reliability using pretrained large model extensions.
\newblock \emph{arXiv preprint arXiv:2207.07411}, 2022.

\bibitem[Van~Amersfoort et~al.(2020)Van~Amersfoort, Smith, Teh, and
  Gal]{van2020uncertainty}
Joost Van~Amersfoort, Lewis Smith, Yee~Whye Teh, and Yarin Gal.
\newblock Uncertainty estimation using a single deep deterministic neural
  network.
\newblock In \emph{International conference on machine learning}, pages
  9690--9700. PMLR, 2020.

\bibitem[van Amersfoort et~al.(2021)van Amersfoort, Smith, Jesson, Key, and
  Gal]{van2021feature}
Joost van Amersfoort, Lewis Smith, Andrew Jesson, Oscar Key, and Yarin Gal.
\newblock On feature collapse and deep kernel learning for single forward pass
  uncertainty.
\newblock \emph{arXiv preprint arXiv:2102.11409}, 2021.

\bibitem[Vedula et~al.(2019)Vedula, Lipka, Maneriker, and
  Parthasarathy]{vedula2019towards}
Nikhita Vedula, Nedim Lipka, Pranav Maneriker, and Srinivasan Parthasarathy.
\newblock Towards open intent discovery for conversational text.
\newblock \emph{arXiv preprint arXiv:1904.08524}, 2019.

\bibitem[Wen et~al.(2016)Wen, Zhang, Li, and Qiao]{wen2016discriminative}
Yandong Wen, Kaipeng Zhang, Zhifeng Li, and Yu~Qiao.
\newblock A discriminative feature learning approach for deep face recognition.
\newblock In \emph{Computer Vision--ECCV 2016: 14th European Conference,
  Amsterdam, The Netherlands, October 11--14, 2016, Proceedings, Part VII 14},
  pages 499--515. Springer, 2016.

\bibitem[Wenzel et~al.(2020)Wenzel, Roth, Veeling, Swiatkowski, Tran, Mandt,
  Snoek, Salimans, Jenatton, and Nowozin]{wenzel2020good}
Florian Wenzel, Kevin Roth, Bastiaan Veeling, Jakub Swiatkowski, Linh Tran,
  Stephan Mandt, Jasper Snoek, Tim Salimans, Rodolphe Jenatton, and Sebastian
  Nowozin.
\newblock How good is the bayes posterior in deep neural networks really?
\newblock In \emph{International Conference on Machine Learning}, pages
  10248--10259. PMLR, 2020.

\bibitem[Winkens et~al.(2020)Winkens, Bunel, Roy, Stanforth, Natarajan, Ledsam,
  MacWilliams, Kohli, Karthikesalingam, Kohl, et~al.]{winkens2020contrastive}
Jim Winkens, Rudy Bunel, Abhijit~Guha Roy, Robert Stanforth, Vivek Natarajan,
  Joseph~R Ledsam, Patricia MacWilliams, Pushmeet Kohli, Alan Karthikesalingam,
  Simon Kohl, et~al.
\newblock Contrastive training for improved out-of-distribution detection.
\newblock \emph{arXiv preprint arXiv:2007.05566}, 2020.

\bibitem[Wu and Goodman(2020)]{wu2020simple}
Mike Wu and Noah Goodman.
\newblock A simple framework for uncertainty in contrastive learning.
\newblock \emph{arXiv preprint arXiv:2010.02038}, 2020.

\bibitem[Yaghoub-Zadeh-Fard et~al.(2020)Yaghoub-Zadeh-Fard, Benatallah, Casati,
  Barukh, and Zamanirad]{yaghoub2020user}
Mohammad-Ali Yaghoub-Zadeh-Fard, Boualem Benatallah, Fabio Casati, Moshe~Chai
  Barukh, and Shayan Zamanirad.
\newblock User utterance acquisition for training task-oriented bots: a review
  of challenges, techniques and opportunities.
\newblock \emph{IEEE Internet Computing}, 24\penalty0 (3):\penalty0 30--38,
  2020.

\bibitem[Zhang(2021)]{zhang2021modern}
Jiaxin Zhang.
\newblock Modern monte carlo methods for efficient uncertainty quantification
  and propagation: A survey.
\newblock \emph{Wiley Interdisciplinary Reviews: Computational Statistics},
  13\penalty0 (5):\penalty0 e1539, 2021.

\bibitem[Zhang et~al.(2024)Zhang, Bi, and Fung]{zhang2024quantification}
Jiaxin Zhang, Sirui Bi, and Victor Fung.
\newblock On the quantification of image reconstruction uncertainty without
  training data.
\newblock In \emph{Proceedings of the IEEE/CVF Winter Conference on
  Applications of Computer Vision}, pages 2072--2081, 2024.

\bibitem[Zheng et~al.(2020)Zheng, Chen, and Huang]{zheng2020out}
Yinhe Zheng, Guanyi Chen, and Minlie Huang.
\newblock Out-of-domain detection for natural language understanding in dialog
  systems.
\newblock \emph{IEEE/ACM Transactions on Audio, Speech, and Language
  Processing}, 28:\penalty0 1198--1209, 2020.

\end{thebibliography}

\end{document}